\documentclass{article}

\usepackage{arxiv}

\usepackage[utf8]{inputenc} 
\usepackage[T1]{fontenc}    
\usepackage{hyperref}       
\usepackage{url}            
\usepackage{booktabs}       
\usepackage{amsfonts}       
\usepackage{nicefrac}       
\usepackage{microtype}      
\usepackage{lipsum}
\usepackage{graphicx}
\usepackage{amsmath}
\usepackage{multirow}
\graphicspath{ {./images/} }

\title{Flow-CDNet: A Novel Network for Detecting Both Slow and Fast Changes in Bitemporal Images}

\author{
Haoxuan Li\thanks{These authors contributed equally to this work.} \\
School of Computer Science and Technology \\
Northwestern Polytechnical University \\
Xi’an 710072, China \\
\texttt{li\_haoxuan@mail.nwpu.edu.cn} \\
\And
Chenxu Wei\footnotemark[1] \\
School of Computer Science and Technology \\
Northwestern Polytechnical University \\
Xi’an 710072, China \\
\texttt{weichenxu@mail.nwpu.edu.cn} \\
\And
Haodong Wang \\
School of Computer Science and Technology \\
Northwestern Polytechnical University \\
Xi’an 710072, China \\
\texttt{traslauc@mail.nwpu.edu.cn} \\
\And
Xiaomeng Hu \\
School of Computer Science and Technology \\
Northwestern Polytechnical University \\
Xi’an 710072, China \\
\texttt{hxm2886@mail.nwpu.edu.cn} \\
\And
Boyuan An \\
School of Computer Science and Technology \\
Northwestern Polytechnical University \\
Xi’an 710072, China \\
\texttt{741099841@mail.nwpu.edu.cn} \\
\And
Lingyan Ran \\
School of Computer Science and Technology \\
Northwestern Polytechnical University \\
Xi’an 710072, China \\
\texttt{lran@nwpu.edu.cn} \\
\And
Baosen Zhang \\
Yellow River Institute of Hydraulic Research \\
Zhengzhou 450003, China \\
\texttt{976129493@qq.com} \\
\And
Jin Jin \\
Yellow River Institute of Hydraulic Research \\
Zhengzhou 450003, China \\
\texttt{jinjin@hky.yrcc.gov.cn} \\
\And
Omirzhan Taukebayev \\
Al-Farabi Kazakh National University \\
Almaty 050040, Republic of Kazakhstan \\
\texttt{omirzhan.taukebayev@kaznu.edu.kz} \\
\And
Amirkhan Temirbayev \\
Al-Farabi Kazakh National University \\
Almaty 050040, Republic of Kazakhstan \\
\texttt{amirkhan.temirbayev@kaznu.edu.kz} \\
\And
Junrui Liu\thanks{Corresponding author: liu.junrui@nwpu.edu.cn} \\
School of Computer Science and Technology \\
Northwestern Polytechnical University \\
Xi’an 710072, China \\
\texttt{liu.junrui@nwpu.edu.cn} \\
\And
Xiuwei Zhang \\
School of Computer Science and Technology \\
Northwestern Polytechnical University \\
Xi’an 710072, China \\
\texttt{xwzhang@nwpu.edu.cn} \\
}

\begin{document}
\maketitle
\begin{abstract}
Change detection typically involves identifying regions with changes between bitemporal images taken at the same location. Besides significant changes, slow changes in bitemporal images are also important in real-life scenarios. For instance, weak changes often serve as precursors to major hazards in scenarios like slopes, dams, and tailings ponds. Therefore, designing a change detection network that simultaneously detects slow and fast changes presents a novel challenge. 
In this paper, to address this challenge, we propose a change detection network named Flow-CDNet, consisting of two branches: optical flow branch and binary change detection branch. The first branch utilizes a pyramid structure to extract displacement changes at multiple scales. The second one combines a ResNet-based network with the optical flow branch's output to generate fast change outputs. Subsequently, to supervise and evaluate this new change detection framework, a self-built change detection dataset Flow-Change, a loss function combining binary tversky loss and L2 norm loss, along with a new evaluation metric called FEPE are designed. Quantitative experiments conducted on Flow-Change dataset demonstrated that our approach outperforms the existing methods. Furthermore, ablation experiments verified that the two branches can promote each other to enhance the detection performance.
\end{abstract}


\section{Introduction}
In real-world monitoring scenarios, both \textbf{slow} and \textbf{fast} changes frequently coexist across various environments. For example, in applications such as slope monitoring, dam safety assessment, and tailings pond management, minor displacements of soil or ore blocks often reflect \emph{slow changes}, while sudden collapses or structural failures correspond to \emph{fast changes}. These phenomena may appear sequentially or simultaneously, thus posing significant challenges for accurate and robust change detection.

Traditional change detection (CD) techniques typically focus on identifying fast changes, where an object appears or disappears completely between two temporal images. These are often tackled using deep neural networks (DNNs) based on semantic segmentation or classification. In contrast, the detection of slow changes---characterized by partial displacement of objects over time---relies on optical flow estimation, which computes pixel-level correspondences between image pairs in the form of dense 2D displacement fields.

In this work, we define the concepts of slow and fast changes more formally: when an object is present in both images but changes its position or shape, we classify it as a \textit{slow change}; if the object exists only in one of the bitemporal images, it is categorized as a \textit{fast change}. This definition is consistent with real-world scenarios and reflects the need for a unified framework capable of simultaneously addressing both change types.

Extensive research has been conducted on the challenges of detecting either slow or fast changes. For slow changes, researchers commonly employ optical flow detection, since it aims to determine the pixel-wise correspondences between source and target images in the form of a 2D displacement field, allowing it to capture minor variations effectively. 
SpyNet~\cite{C2} employs a coarse-to-fine approach that combines traditional methods with deep learning techniques. 
ContinuousFlow~\cite{C4} combines occlusion and cost volume techniques with optical flow estimation. 
MaskFlownet~\cite{C5} introduces a learnable occlusion mask in the asymmetric feature matching module, thus improving the effectiveness of optical flow prediction.
LiteFlowNet~\cite{C3} uses traditional brightness inconsistency mapping to address occlusion issues.
RAFT~\cite{C6} extracts per-pixel features and builds multi-scale 4D cost volumes for all pixel pairs. It maintains and updates a single fixed-resolution optical flow image with high resolution, so it has strong cross-dataset generalization ability. 
AccFlow~\cite{wu2023accflow} accumulates frame-to-frame optical flow to obtain long-distance cross-frame optical flow, adapting to optical flow estimation algorithms for arbitrary frame pairs. VideoFlow~\cite{shi2023videoflow} can thoroughly explore and utilize multi-frame data, significantly enhancing the performance of optical flow estimation.
For fast changes, researchers commonly employ change detection with deep neural networks. 
PSPNet~\cite{C9} is the first to introduce the concept of pyramid pooling modules and integrate global contextual information. Due to its ability to leverage global contextual information through context aggregation from different regions, it has become a baseline method in the realm of deep change detection. 
Chen et al.~\cite{C19} introduces a network model called DASNet for high-resolution image change detection. 
Liu et al.~\cite{C10} utilizes semantic segmentation as an auxiliary task to aid change detection. This approach helps to learn more distinctive object-level features, thereby enhancing the quality of change detection results. 

Despite recent advancements, there remains a critical limitation: existing methods cannot effectively detect slow and fast changes simultaneously. Optical flow models struggle with occlusions and object disappearance, while conventional CD networks are insufficiently sensitive to subtle, continuous transformations.

To overcome this limitation, we propose a novel framework named \textbf{Flow-CDNet}. This dual-branch architecture integrates optical flow estimation and change detection in a unified learning paradigm. Specifically, one branch estimates dense motion maps using a pyramid-based optical flow module, while the other branch performs binary change detection by leveraging both the original bitemporal images and the motion features. The joint learning mechanism enables mutual enhancement between the two branches, improving the network's capability to detect diverse change types with higher accuracy.

To facilitate model training and evaluation, we construct a dedicated dataset named \textbf{Flow-Change}, which includes synthetic bitemporal image pairs with annotated labels for both optical flow and binary change detection. Additionally, we design a composite loss function that jointly optimizes optical flow regression and change detection objectives. Furthermore, we introduce a new evaluation metric called \textbf{FEPE} (F1-score over End-Point Error), which provides a unified assessment of model performance across both slow and fast changes.
Finally, we validate the generalization capability of Flow-CDNet on real-world dam bank images, demonstrating its effectiveness in practical applications.

The main contributions of this work are summarized as follows:

\begin{itemize}
    \item We propose \textbf{Flow-CDNet}, a unified change detection framework that combines pyramid-based optical flow estimation with residual convolutional change classification in a dual-branch architecture, enabling simultaneous detection of both slow and fast changes in bitemporal imagery.
    \item We construct a new dataset named \textbf{Flow-Change} and propose a \textbf{composite loss function} and a novel evaluation metric (\textbf{FEPE}) to support end-to-end training and performance assessment.
    \item We assess the proposed approach on both real-world dam bank monitoring scenarios and the constructed synthetic dataset, where it effectively identifies both abrupt collapses and gradual deformations. These results, visualized in detail, not only demonstrate strong practical applicability but also highlight the complementary strengths of the dual-branch architecture through ablation studies.
\end{itemize}

\section{Related Works}

\subsection{Optical Flow Estimation}

Optical flow estimation techniques are broadly categorized into traditional and deep learning-based methods. Conventional approaches such as EpicFlow~\cite{C13}, DeepFlow~\cite{C25}, and MirrorFlow~\cite{C26} propose various strategies to enhance estimation accuracy, yet they generally fall short in meeting real-time performance requirements.

The advent of deep learning revolutionizes the field. FlowNet~\cite{C15} pioneers the use of convolutional neural networks for optical flow, achieving real-time operation but with limited precision. FlowNet2.0~\cite{C37} addresses this shortcoming by introducing a stacked architecture, utilizing synthetic datasets, and adopting improved training protocols, leading to notable accuracy gains. SpyNet~\cite{C2} introduces a spatial pyramid structure with coarse-to-fine warping and residual prediction, enabling efficient large displacement flow estimation with low computational cost. LiteFlowNet3~\cite{C38} further refines performance by handling outliers in the cost volume through adaptive modulation prior to decoding and correcting distorted flows using nearby reliable estimates.

To better handle large displacement motion, GMFlowNet~\cite{C20} and GMFlow~\cite{C21} reformulate optical flow as a matching problem rather than a regression one, enhancing robustness and accuracy. Jiang et al.~\cite{C16} show that precise flow estimation is attainable even when matching only a small subset of pixels, emphasizing the efficiency of sparse correspondences.

Transformer-based architectures also emerge as a powerful paradigm. FlowFormer~\cite{C27} introduces transformers into this domain, employing a cost volume encoder for compact representation and a recursive decoder that iteratively refines flow using dynamic location queries. SAMFlow~\cite{C40} extends this by integrating a frozen SAM image encoder and improving semantic perception of objects within the scene.

Robustness across scales and detail preservation are addressed by approaches like AnyFlow~\cite{C31}, which estimates flow from images of varying resolutions, excelling at capturing fine-grained motion. DistractFlow~\cite{C32} contributes a novel data augmentation method by blending optical flow estimates with structurally similar disturbance images, aligning visual perturbations with real-world appearances.

In pursuit of greater efficiency, RAPIDFlow~\cite{C39} incorporates NeXt1D convolutional blocks within a fully recursive feature pyramid, reducing computational load without compromising accuracy. MaxFlow~\cite{C41} adopts 1D matching along with MaxViT transformers to significantly lower complexity while retaining strong performance. MatchFlow~\cite{C42} enhances generalization through geometric pretraining and employs a QuadTree attention mechanism to boost adaptability across datasets.

Innovative attention-based designs continue to improve correlation modeling. CRAFT~\cite{C43} revitalizes the correlation volume using a cross-attentional transformer, achieving resilience against blur and substantial motion. KPA-Flow~\cite{C44} introduces kernel patch attention to strengthen local context modeling, setting new performance records on benchmarks like Sintel and KITTI.

Targeted solutions address specific limitations. I-RAFT~\cite{C45} replaces zero-initialized flow with a multi-scale initialization strategy, improving accuracy while reducing model size. FlowDiffuser~\cite{C58} reconceptualizes flow as a conditional generation problem using diffusion models, leveraging a noise-to-flow mechanism with a Conditional Recurrent Denoising Decoder. DeepPyNet~\cite{C59} delivers a lightweight feature pyramid and a 4D correlation structure, achieving efficient performance with a minimal parameter footprint. PatchFlow~\cite{C60} enables high-resolution flow estimation on resource-constrained devices through a two-stage patch-based method.

Novel theoretical frameworks also improve the modeling of optical flow. DEQFlow~\cite{C61} formulates optical flow estimation as an equilibrium computation, where the solution corresponds to the fixed point of an implicit iterative process modeled by a deep equilibrium network. This formulation enables constant memory usage during training and inference, while also enhancing convergence stability. Equivariant Flow~\cite{C62} introduces equivariant neural architectures to mitigate the inherent direction-dependent biases in conventional models, thereby improving generalization and robustness under varying motion dynamics.

Recent developments in remote sensing have introduced change detection networks that share foundational principles with optical flow, such as pixel-wise semantic differentiation and temporal consistency. Zhang et al.~\cite{zhang2024remote} propose a semi-supervised contrastive learning approach for semantic segmentation to enhance change detection, emphasizing effective feature discrimination even with limited labeled data. DifUnet++~\cite{zhang2021difunet++} combines UNet++ with a differential pyramid module to better capture hierarchical spatial differences in satellite imagery. ADHR-CDNet~\cite{zhang2022adhr} introduces attentive mechanisms to refine high-resolution spatial features, enabling precise detection of localized changes. These models illustrate the efficacy of segmentation-driven feature extraction, which complements optical flow estimation in temporally varying scenes.

Among all deep learning methods, RAFT~\cite{C6} stands out as a foundational model that redefined the optical flow landscape through its iterative refinement strategy and all-pairs correlation volume. RAFT achieves a rare balance between accuracy and generalization by jointly encoding local and global motion cues while maintaining computational feasibility. Its elegant architecture and impressive benchmark performance have made it a cornerstone for many subsequent innovations, and it serves as the backbone for our proposed work.

\subsection{Change Detection}
Before the advent of deep learning, traditional change detection techniques play a pivotal role in remote sensing and computer vision. Among these, Principal Component Analysis (PCA) and Change Vector Analysis (CVA) are extensively employed in multispectral and hyperspectral imagery to reduce data dimensionality and accentuate spectral variations indicative of change. Post-classification comparison methods, which involve independently classifying each temporal image and subsequently contrasting the outputs, are widely utilized for thematic change mapping. Additionally, unsupervised clustering approaches such as K-Means and ISODATA are applied to difference images, enabling the identification of change regions without reliance on labeled data.

In recent years, the rapid progression of deep learning has revolutionized the landscape of change detection, delivering significant improvements in both accuracy and efficiency~\cite{C8}~\cite{C36}. Long et al.~\cite{C22} pioneer the use of Fully Convolutional Networks (FCNs) for end-to-end semantic segmentation, laying the groundwork for numerous deep learning-based change detection frameworks. Building upon this paradigm, Dault et al.~\cite{C18} propose three FCN-based architectures tailored to change detection tasks. Tezcan et al.~\cite{C17} introduce BSUVNet, a supervised model leveraging FCNs for occlusion-aware background subtraction in video sequences, with auxiliary semantic segmentation inputs to enhance detection fidelity.

To address the challenges of dynamic scene analysis, Ou et al.~\cite{C33} present the Deep Frame Difference Convolutional Neural Network (DFDCNN), which incorporates two specialized subnetworks—DifferenceNet and AppearanceNet—that collaboratively predict foreground segmentation maps. Cheng et al.~\cite{C34} develop a multi-flow convolutional framework for action recognition in video data, employing sparse temporal sampling, frame differencing, and multi-branch learning to extract rich spatiotemporal features. Zheng et al.~\cite{C35} propose a hybrid architecture combining convolutional layers with transformer modules, designed to extract both global and local change features through a tri-branch structure enhanced by spatial and channel-level dual attention mechanisms. Parallel research efforts also target heterogeneous change detection problems~\cite{xing}.

Yang et al.~\cite{C46} introduce DLCDet, a dictionary learning-based framework integrated with a feature pyramid network and dual supervision strategy, effectively addressing semantic inconsistencies across temporal scenes and improving detection under seasonal variation. Zhang et al.~\cite{C47} propose a deep Siamese network augmented with a contextual transformer module, employing multi-scale fusion and self-attention to strengthen bi-temporal feature alignment and pixel-wise precision. Sun et al.~\cite{C48} devise a hybrid model combining Convolutional Neural Networks (CNNs) and Graph Neural Networks (GNNs), enhanced with hierarchical feature supervision to facilitate fine-grained semantic interactions in high-resolution change detection.

Wang et al.~\cite{C49} design a building change detection pipeline that synergizes pixel-level and object-level cues to generate saliency-guided difference maps, further refined through fuzzy clustering and a deep classification network. In low-data regimes, Paul et al.~\cite{C50} explore a transfer learning strategy utilizing a pre-trained VGG19 backbone and a lightweight FCN, followed by SVM classification, yielding improved generalization. Nie et al.~\cite{C51} develop a semi-supervised framework integrating Generative Adversarial Networks (GANs), residual Siamese networks, flow alignment, and atrous convolutions to robustly handle limited labeled data scenarios in remote sensing.

Zhu et al.~\cite{C52} propose ChangeViT, a Vision Transformer-based architecture featuring a detail-capturing mechanism and feature injection module, achieving state-of-the-art performance on large-scale fine-grained benchmarks. Dong et al.~\cite{C53} formulate a multimodal change detection approach that integrates image-text embeddings from CLIP, supplemented by a differential compensation module to enhance semantic change localization. Yin et al.~\cite{C54} present a vision-language joint learning framework for box-supervised change detection, leveraging transformer-based fusion of visual and textual modalities.

Chen et al.~\cite{C55} introduce SGANet, a geometry-aware Siamese network incorporating RGB imagery and monocular depth estimation, guided by cross-attention mechanisms to improve spatial localization through geometric cues. Meng et al.~\cite{C56} propose ChangeAD, which combines bi-temporal alignment with differential feature integration to enhance robustness against seasonal and illumination variations. Finally, Fazry et al.~\cite{C57} develop LocalCD, a locality-sensitive Vision Transformer that substitutes conventional feed-forward layers with depth-wise convolutions, enabling more accurate delineation of change boundaries.

\section{Proposed Method}
\subsection{Network Architecture}
\begin{figure}[!htb]
    \centering
    \includegraphics[width=0.8\linewidth]{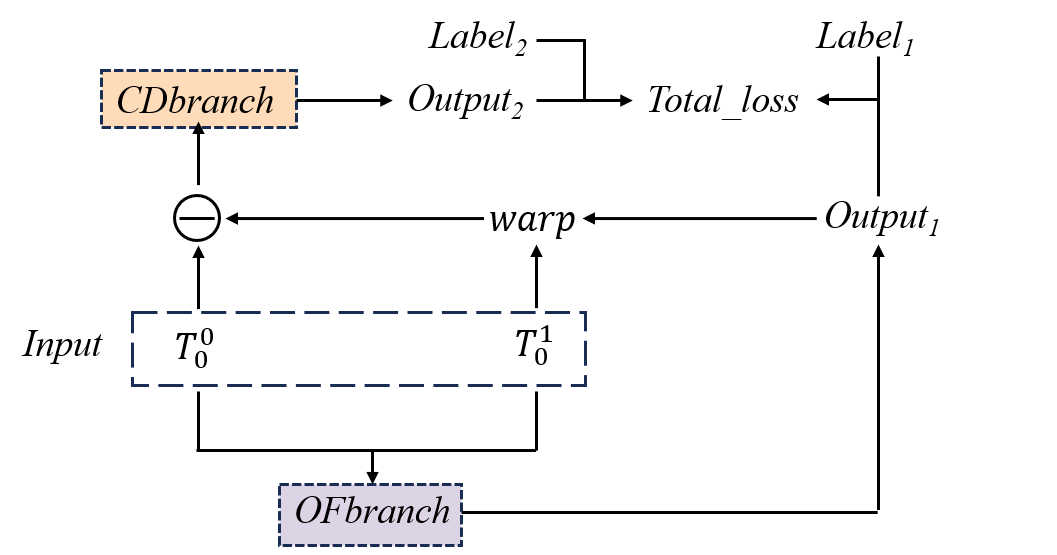}
    \caption{Overview of the proposed Flow-CDNet framework.}
    \label{fig:flow-CDNet}
\end{figure}

In this section, we build up a framework named Flow-CDNet for simultaneous slow and fast change detection. As shown in Fig.~\ref{fig:flow-CDNet}, it consists of an optical flow detection branch (OFbranch) and a classical change detection branch (CDbranch). The OFbranch  employs a pyramid structure to extract displacement changes at multiple scales, as depicted in Fig.~\ref{fig:OFbranch}. The CDbranch utilizes a network architecture based on spatial pyramid pooling to transform the output results into binary images with CNN, illustrated in Fig.~\ref{fig:CDbranch}. The following subsections provide details.

\subsubsection{Structure of Flow-CDNet.}
The proposed Flow-CDNet model is designed to jointly learn optical flow estimation and change detection from bitemporal image pairs, thereby enabling accurate identification of both fast and slow scene changes. Given a pair of temporally separated images $T^0_{0}$ and $T^1_{0}$, which represent the same scene captured at two different time points, the network performs a sequence of processing steps to extract motion and change-related information through two interdependent branches.

Firstly, the input image pair is forwarded through the Optical Flow Estimation Branch, which is responsible for estimating the pixel-wise motion between $T^0_{0}$ and $T^1_{0}$. This branch outputs a dense motion estimation map, denoted as $output_1$, representing the displacement of each pixel from one image to the other.

Following this, the image $T^1_{0}$ is warped according to the estimated flow field $output_1$, resulting in a motion-aligned image. This operation aims to minimize the misalignment between the two images due to object or scene motion, thus allowing more accurate identification of residual differences attributable to genuine scene changes.

Subsequently, the absolute pixel-wise difference between the original image $T^0_{0}$ and the motion-compensated version of $T^1_{0}$ is computed. This difference map serves as a preliminary indicator of potential changes, though it may still contain noise or artifacts, especially in regions with inaccurate motion estimation.

To address this, the model incorporates an adaptive mask mechanism designed to emphasize regions likely to exhibit fast or abrupt changes. This mechanism adaptively weights the features based on motion uncertainty and intensity, thereby enhancing the network's ability to localize meaningful change patterns while suppressing false positives from low-confidence areas.

The refined difference features, together with the flow estimation $output_1$, are then fed into the Change Detection Branch. This second branch processes the combined information and produces a binary or probabilistic change map, denoted as $output_2$, which reflects the spatial distribution of detected changes between the two input frames.

By explicitly using $output_1$ as an auxiliary input to the CDbranch, the network effectively integrates both motion and appearance cues, forming a tightly coupled dual-branch architecture. This integrated design—central to the Flow-CDNet framework—has been empirically validated through ablation studies, which demonstrate that the inclusion of optical flow guidance significantly enhances the precision and robustness of change detection, particularly in the presence of dynamic background motion.

Finally, the overall training objective of the network is formulated as a multi-task loss function, which jointly optimizes both branches. The total loss aggregates the error terms from the flow estimation branch (related to $output_1$) and the change detection branch (related to $output_2$), thereby encouraging mutual reinforcement between the two tasks during end-to-end training.

\subsubsection{Optical Flow Detection Branch.}
\begin{figure}[!htb]
    \centering
    \includegraphics[width=0.8\linewidth]{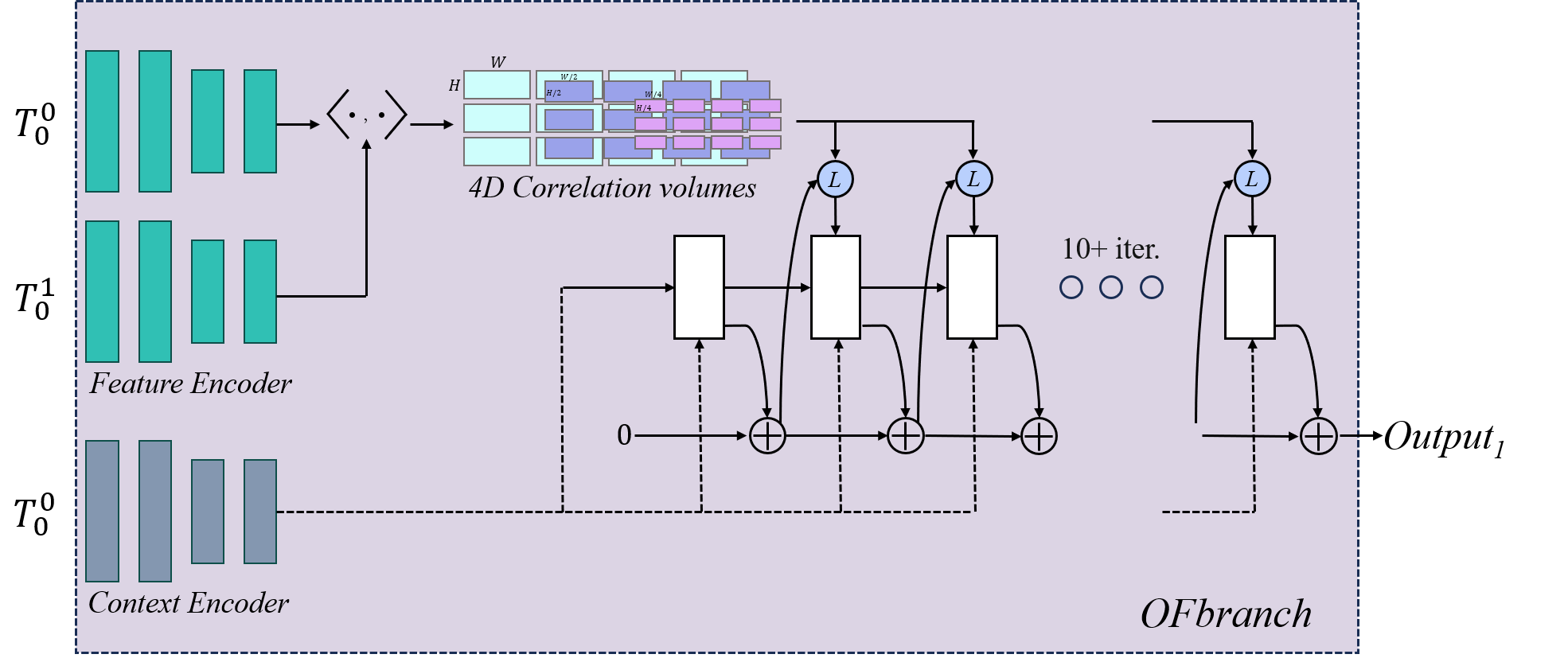}
    \caption{OFbranch.}
    \label{fig:OFbranch}
\end{figure}

As shown in Fig.~\ref{fig:OFbranch}, the input to the OFbranch is a pair of bitemporal images, i.e., $T^0_0$ and $T^1_0$ with the size of ($H,W$). The objective is to estimate a dense displacement field $\mathbf{f} = (f^1, f^2) \in \mathbb{R}^{H \times W \times 2} $, which maps each pixel $(u, v)$ in $T^0_0$ to its corresponding coordinates $(u', v') = (u + f^1(u), v + f^2(v))$ in $T^1_0$. In our model, the OFbranch is implemented using the RAFT (Recurrent All-Pairs Field Transforms) architecture ~\cite{C6}, which achieves state-of-the-art performance in optical flow estimation through iterative updates based on a high-resolution correlation volume and learned update operators. OFbranch consists of 3 main components,  feature extraction, computing visual similarity and iterative updates.

For feature extraction, The feature encoder $ g_\theta $, implemented as a convolutional network with residual blocks, extracts high-dimensional features from both input images. It progressively downsamples the spatial resolution to $ 1/8 $ of the original dimensions, generating feature maps $ g_\theta(T_0^0) $ and $ g_\theta(T_0^1) \in \mathbb{R}^{H/8 \times W/8 \times 256} $. A separate context encoder $ h_\theta $, sharing the same architecture as $ g_\theta $, processes only $ T_0^0 $ to produce a contextual feature map $ h_\theta(T_0^0) $. This contextual information encodes semantic priors to guide motion boundary refinement during iterative updates.

For visual Similarity Computation, visual similarity between $T_0^0 $ and $T_0^1 $ is quantified by computing the inner product of all feature pairs across the two images. The resulting 4D correlation volume $C \in \mathbb{R}^{H/8 \times W/8 \times H/8 \times W/8} $ is defined as: 

\begin{equation}
C_{ijkl} = \sum_{h=1}^{256} g_\theta(T_0^0)_{ijh} \cdot g_\theta(T_0^1)_{klh},
\end{equation}

where $(i,j) $ and $(k,l) $ denote spatial positions in $T_0^0 $ and $T_0^1 $, respectively. To capture both large and small displacements, a multi-scale pyramid $\{C^1, C^2, C^3, C^4\} $ is constructed by applying average pooling to the last two dimensions of $C $. Pooling kernel sizes $\{1, 2, 4, 8\} $ correspond to progressively coarser resolutions, enabling hierarchical matching across varying displacement ranges.

For iterative updates, The optical flow field $\mathbf{f} $ is iteratively refined from an initial estimate $ \mathbf{f}_0 = \mathbf{0}$ using a lightweight convolutional GRU unit. At each iteration $ k $:

(a) For each pixel $ \mathbf{x} = (u,v) $ in $ T_0^0 $, the current flow estimate $ \mathbf{f}_k $ maps $ \mathbf{x} $ to a correspondence $ \mathbf{x}' = (u + f^1_k(u), v + f^2_k(v)) $ in $ T_0^1 $. A local grid $ \mathcal{N}(\mathbf{x}')_r = \{\mathbf{x}' + \mathbf{dx} \, | \, \|\mathbf{dx}\|_1 \leq r\} $ is defined around $ \mathbf{x}' $, and multi-scale correlation features are retrieved via bilinear interpolation from the pyramid $ \{C^1, C^2, C^3, C^4\} $. 

(b) The retrieved correlation features, flow features (encoded from $ \mathbf{f}_k $), and contextual features $ h_\theta(T_0^0) $ are concatenated into a unified input tensor. 

(c) The input tensor is processed by a convolutional GRU cell to predict a flow increment $ \Delta \mathbf{f} $. The hidden state $ \mathbf{h}_t $ and flow update are governed by: 

\begin{equation}
\begin{aligned}
\mathbf{z}_t &= \sigma\left(\text{Conv}_{3\times3}([\mathbf{h}_{t-1}, \mathbf{x}_t], \mathbf{W}_z)\right), \\
\mathbf{r}_t &= \sigma\left(\text{Conv}_{3\times3}([\mathbf{h}_{t-1}, \mathbf{x}_t], \mathbf{W}_r)\right), \\
\tilde{\mathbf{h}}_t &= \tanh\left(\text{Conv}_{3\times3}([\mathbf{r}_t \odot \mathbf{h}_{t-1}, \mathbf{x}_t], \mathbf{W}_h)\right), \\
\mathbf{h}_t &= (1 - \mathbf{z}_t) \odot \mathbf{h}_{t-1} + \mathbf{z}_t \odot \tilde{\mathbf{h}}_t,
\end{aligned}
\end{equation}

where $ \mathbf{x}_t $ denotes the fused features, and $ \odot $ represents element-wise multiplication. The flow field is updated as $ \mathbf{f}_{k+1} = \mathbf{f}_k + \Delta \mathbf{f} $. This recurrent process, with shared weights across iterations, enables stable convergence even after 100+ updates, avoiding error propagation inherent in coarse-to-fine approaches.

The final low-resolution flow field $ \mathbf{f}_N \in \mathbb{R}^{H/8 \times W/8 \times 2} $ is upsampled to full resolution $ output_1 \in \mathbb{R}^{H \times W \times 2} $ using a convex combination strategy. 

\subsubsection{Change Detection Branch.}
\begin{figure}[!htb]
    \centering
    \includegraphics[width=0.8\linewidth]{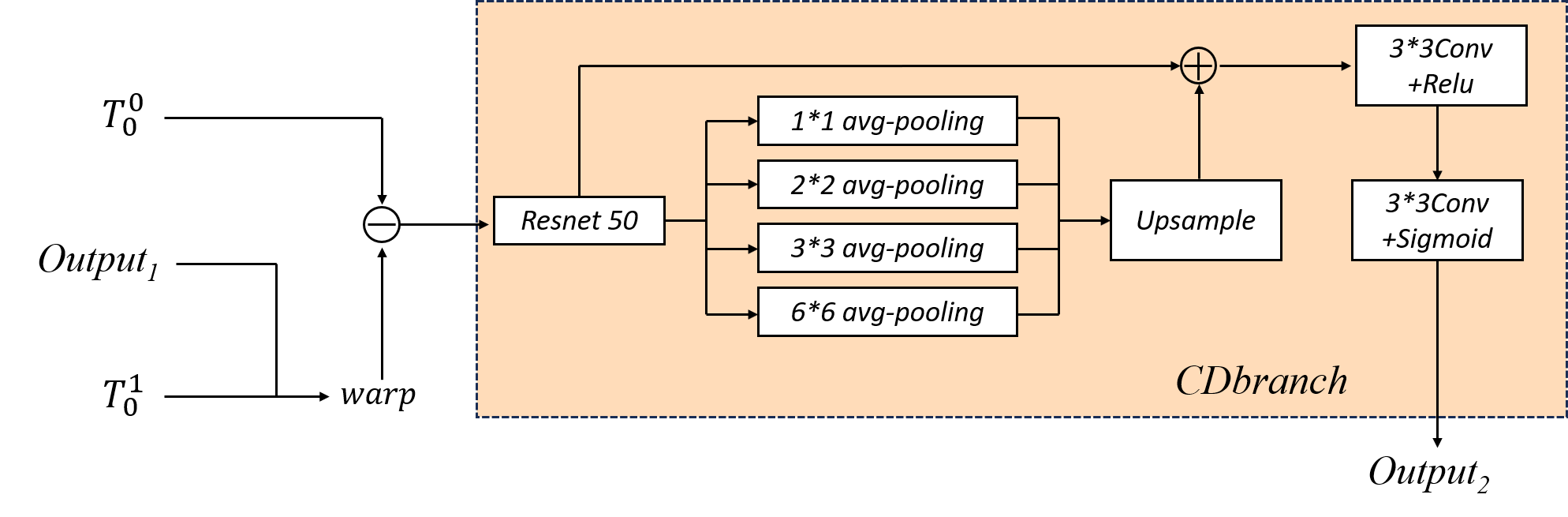}
    \caption{CDbranch.}
    \label{fig:CDbranch}
\end{figure}

As shown in Fig.~\ref{fig:CDbranch}, the CDbranch has three inputs: the input image $T^0_0$ and $T^1_0$, and $output_1$ obtained from the OFbranch. First, the image $T^1_0$ is warped using the predicted optical flow $output_1$ to generate a motion-compensated image, denoted as $w(T^1_0,output_1)$, then compute the absolute difference between $T^0_0$  and $w(T^1_0,output_1)$. Subsequently, a mask mechanism employing optical flow magnitude analysis dynamically identifies slow-changing regions, effectively suppressing their interference in the change detection process. Second, input the absolute difference result into the convolutional block using ResNet50 as backbone, and the output is denoted as $F_0$. Third, input $F_0$ into four parallel average pooling operations to generate four feature maps with different sizes, noted as $F_1,F_2,F_3,F_4$. Fourth, upsample these four feature maps to the same size as $F_0$, and perform channel stacking with the original feature map $F_0$,  then pass the stacked feature maps through a $3\times3$ convolution, regularization and ReLU to obtain the feature map $F_5$ with 512 channels. Fifth, through $3\times3$ convolution and Sigmoid, a single-channel binary change feature map $F_6$ is obtained as the final output of the CDbranch, denoted as $output_2$.

\subsection{Loss Function}
To evaluate the results of optical flow detection, it is necessary to remove the binary changed regions ($label_2$) and extract only the regions containing slow change, then calculate the L2 norm loss with $label_1$ and $output_1$, as shown in Equation \eqref{equation_lossl2}.
\begin{equation}
loss_{l2}=||output_1-label_1 ||_2\cdot (1-label_2)
\label{equation_lossl2}
\end{equation}
To evaluate the results of change detection, we employ the change detection labels ($label_2$) and $output_2$ to compute the Tversky Loss\cite{tversky}, outlined in Equation \eqref{equation_losstversky}.

\begin{equation}
\begin{aligned} 
loss_{Tversky}=\frac{masked\_gt}{masked\_gt+\alpha\cdot wrong\_classified + \beta \cdot unmasked\_gt}
\label{equation_losstversky}
\end{aligned}
\end{equation}
Where $masked\_gt=output_2\cdot label_2$, $unmasked\_gt=output_2\cdot(1-label_2)$, $wrong\_classified=(1-output_2)\cdot label_2$, $\alpha$ and $\beta$ are hyperparameters that control the penalty weights for $wrong\_classified$ and $unmasked\_gt$, respectively. 
 By considering the distinct magnitudes of the two losses, we assign a weight $\psi$ to Tversky loss. The overall training loss is presented in Equation \eqref{equation_losstotal}.  
\begin{equation}
loss_{total}=loss_{l2}+\psi \cdot loss_{Tversky}
\label{equation_losstotal}
\end{equation}

\subsection{Evaluation Metric}

The widely used evaluation criterion, F1-score and EPE (End-Point Error) are adopted to evaluate the performance of binary change detection and optical flow. The F1-score is the statistical analysis of predictions, considering the value of true positive (TP), false positive (FP), and false negative (FN). They are defined as:
\begin{equation}
F_1=2/(Precision^{-1}+Recall^{-1})
\end{equation}
\begin{equation}
Precision={TP}/(TP+FP)
\end{equation}
\begin{equation}
Recall={TP}/(TP+FN)
\end{equation}

The EPE metric calculates the euclidean distance between the estimated optical flow$(\vec{F})$ and the ground truth $(\vec{F_{gt}})$, as defined in Equation \eqref{equation_epe}:
\begin{equation}
EPE=\sqrt{|\vec{F}-\vec{F_{gt}}|}
\label{equation_epe}
\end{equation}
Considering regions with subtle motion, the mean EPE (mEPE) metric is adopted, which calculates the average error over all pixels within the union of regions with offset in the ground truth labels (denoted as $Q$) and regions with offset in the predicted outputs (denoted as $Q^*$), as described in Equation \eqref{equation_mepe}:
\begin{equation}
mEPE=\sum_{\substack{i\in Q\cup Q^*}} \frac{\sqrt{|\vec{F_{gt}^i}-\vec{F^i}|}}{||Q\cup Q^*||}
\label{equation_mepe}
\end{equation}

While, to simultaneously evaluate binary change detection and optical flow results, we need a comprehensive evaluation indicator, so a new evaluation criterion namely FEPE is designed to combine the F1-score and the mEPE metric with $\epsilon$ being a low perturbation, as shown in Equation \eqref{equation_fepe}.

\begin{equation}
FEPE=\frac{F_1}{mEPE+\epsilon}
\label{equation_fepe}
\end{equation}

When F1-score becomes larger/smaller (indicating better/worse change detection performance) and mEPE metric becomes smaller/larger (indicating better/worse optical flow estimation performance), FEPE metric tends to become larger/smaller (indicating better/worse prediction performance). This design allows for a comprehensive evaluation that considers both change detection and optical flow estimation.

\section{Experiments}
\subsection{Synthetic Flow-Change Dataset}
Due to the relatively low occurrence of hazards such as deformations in dam bank and other slope areas, there is a scarcity of preserved image data, which makes it challenging to provide sufficient training data for deep models. Additionally, there is currently no existing change detection dataset that includes both fast and slow changes simultaneously. To train and evaluate the proposed method, a new synthetic dataset named Flow-Change is built by integrating FlyingChairs dataset \cite{C15} (an optical flow dataset) with PASCAL VOC 2007 dataset \cite{2015} (a semantic segmentation dataset). 

\begin{figure}[!htb]
    \centering
    \includegraphics[width=0.8\linewidth]{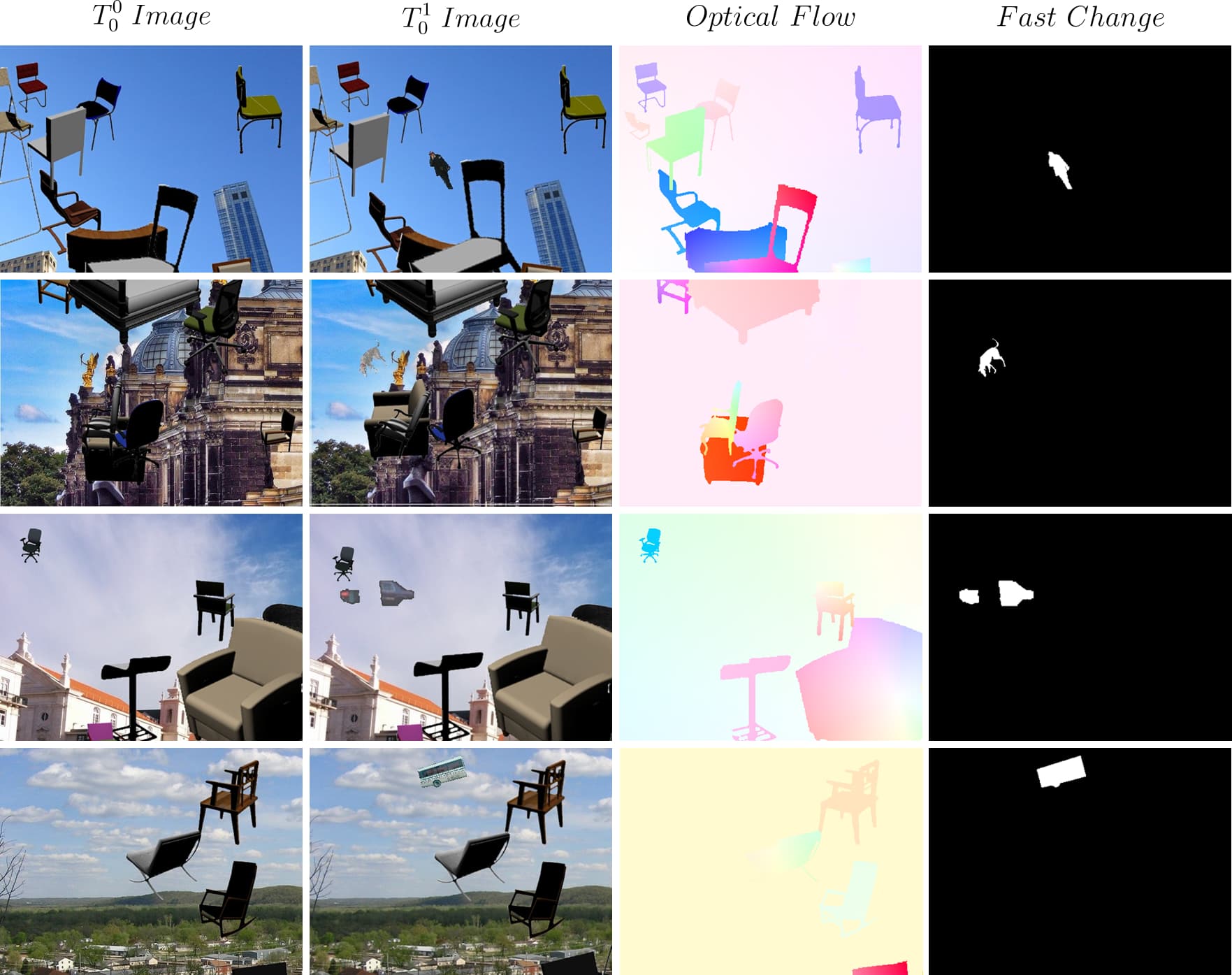}
    \caption{Dataset Visualization, $T_0^0$ represents image in the first frame, $T_0^1$ represents image in the second frame, Optical Flow column shows the ground truth transition between frame one and two, and Fast Change column shows ground truth of the item that only showed up in frame two.}
    \label{fig:dataset_visualize}
\end{figure}

As shown in Fig.~\ref{fig:dataset_visualize}, each data instance comprises a total of 4 images with size of $512\times384$ pixels, including a bitemporal image pair synthetized from FlyingChairs and PASCAL VOC, optical flow detection label map, and change detection label map. The training set consists of 11,736 pairs of images, while the test set includes 2,935 pairs of images. 

The specific process for creating this dataset is as follows: each pair of bitemporal images is sequentially selected from the FlyingChairs dataset as the background. Then, potential change object regions are extracted from the PASCAL VOC 2007 dataset (e.g., humans, animals, vehicles), and subjected to random transformations including scaling, rotation, and random channel shuffle, then these objects are pasted onto the images to simulate fast change scenarios. Additionally, typical data augmentation techniques such as image brightness and contrast enhancement are applied to enrich the synthesized dataset. The corresponding optical flow detection labels from FlyingChairs and binary change detection labels, computed based on the pasting positions, are recorded as the detection ground truth.

\subsection{Experiment on Flow-Change Dataset}

\textbf{Experiment settings.} Our experiments are conducted on a GPU cluster comprising four NVIDIA GeForce GTX 4090 accelerators. The network architecture implements distinct learning rate configurations: the optical flow (OF) branch is initialized with 1e-5 learning rate, while the change detection (CD) branch employs 1e-4. We configure the optimization framework with AdamW and train for 1,000 epochs using batch size 4. We adopt the Tversky Loss for the CD branch, with $\alpha$ set to 0.7 and $\beta$ set to 0.3. The multi-task loss weighting coefficient $\phi$ maintains a fixed ratio of 10 throughout all experiments. 

\textbf{Comparison experiments.} The quantitative comparison results on Flow-Change dataset are displayed in Table \ref{tab:sota}. Since there is no similar network for both fast and slow change detection, different backbones of optical flow estimation are adopted to construct Flow-CDNet like networks. Flow-CDNet-L utilizes LiteFlowNet\cite{C3} and CDNet as backbone, Flow-CDNet-R utilizes RAFT\cite{C6} and CDNet as backbone, Flow-CDNet utilizes SpyNet{C2} and CDNet as backbone. The best-performing results are shown in bold, and the second-best results are underlined. As shown in Table 1, it is evident that Flow-CDNet outperforms all compared backbones and achieves the highest FEPE metric of 0.869. Compared to the second-best optical flow detection backbone RAFT, our method achieves an EPE improvement of 1.382 and a 3.2\% higher F1-score than the next best backbone (SpyNet+CDNet).

\begin{table}
\centering

\begin{tabular}{lccc}
    \toprule
    Methods& F1-score$\uparrow$& mEPE$\downarrow$& FEPE$\uparrow$\\
    \toprule
    CDNet\cite{C9}& 0.753& -& - \\
    SpyNet\cite{C2}& - & 3.383& - \\
    LiteFlowNet\cite{C3}& - & 6.433& - \\
    RAFT\cite{C6}& - & \underline{2.409}& - \\
    Flow-CDNet-L(LiteFlowNet+CDNet)& 0.821 & 5.720& 0.144 \\
    Flow-CDNet-S(SpyNet+CDNet)&\underline{0.860} &2.798&\underline{0.308} \\
    Flow-CDNet&\textbf{0.892} & \textbf{1.027}& \textbf{0.869}\\
    \toprule
    \end{tabular}
\caption{Quantitative comparison results with the state-of-the-art methods on the Flow-Change dataset. The best in bold, and the second-best is underlined.}
\label{tab:sota}
\end{table}

\textbf{Ablation Study.} Ablation experiments are conducted on Flow-Change dataset to verify the effectiveness of the OFbranch and the CDbranch, and further to analyze the impact of different OFbranch design on network performance.  

Table~\ref{tab:ablation} displays the ablation study results. Compared with only using the OFbranch, Flow-CDNet achieves a higher mEPE metric, indicating that the result ($output_2$) of change detection has a positive contribution to the overall detection results. In contrast, compared with only using the change detection branch, Flow-CDNet exhibits a higher F1-score. This indicates that the result ($output_1$) of optical flow detection also contributes to the overall detection results. This experiment demonstrates that the two branches can mutually enhance each other, improving Flow-CDNet's ability to detect both slow and fast changes.

\begin{table}
\centering
\begin{tabular}{ccccc}
    \toprule
    Branch &  &Metric& &\\
    \cmidrule(r){1-2}\cmidrule(r){3-5}
    OFbranch & CDbranch&  F1-score$\uparrow$& mEPE$\downarrow$& FEPE$\uparrow$ \\
    \toprule
    $\surd$ & - & - & 2.409 & - \\
    - & $\surd$ & 0.753 & - & - \\
    $\surd$ & $\surd$ & \textbf{0.892} & \textbf{1.027} & \textbf{0.869} \\
    \toprule
\end{tabular}
\caption{Ablation study on selecting different branches. "$\surd$" indicates that the network utilizes the corresponding branch. The best-performing result is in bold.}
\label{tab:ablation}
\end{table}

\begin{figure}[!ht]
    \centering
    \includegraphics[width=\textwidth]{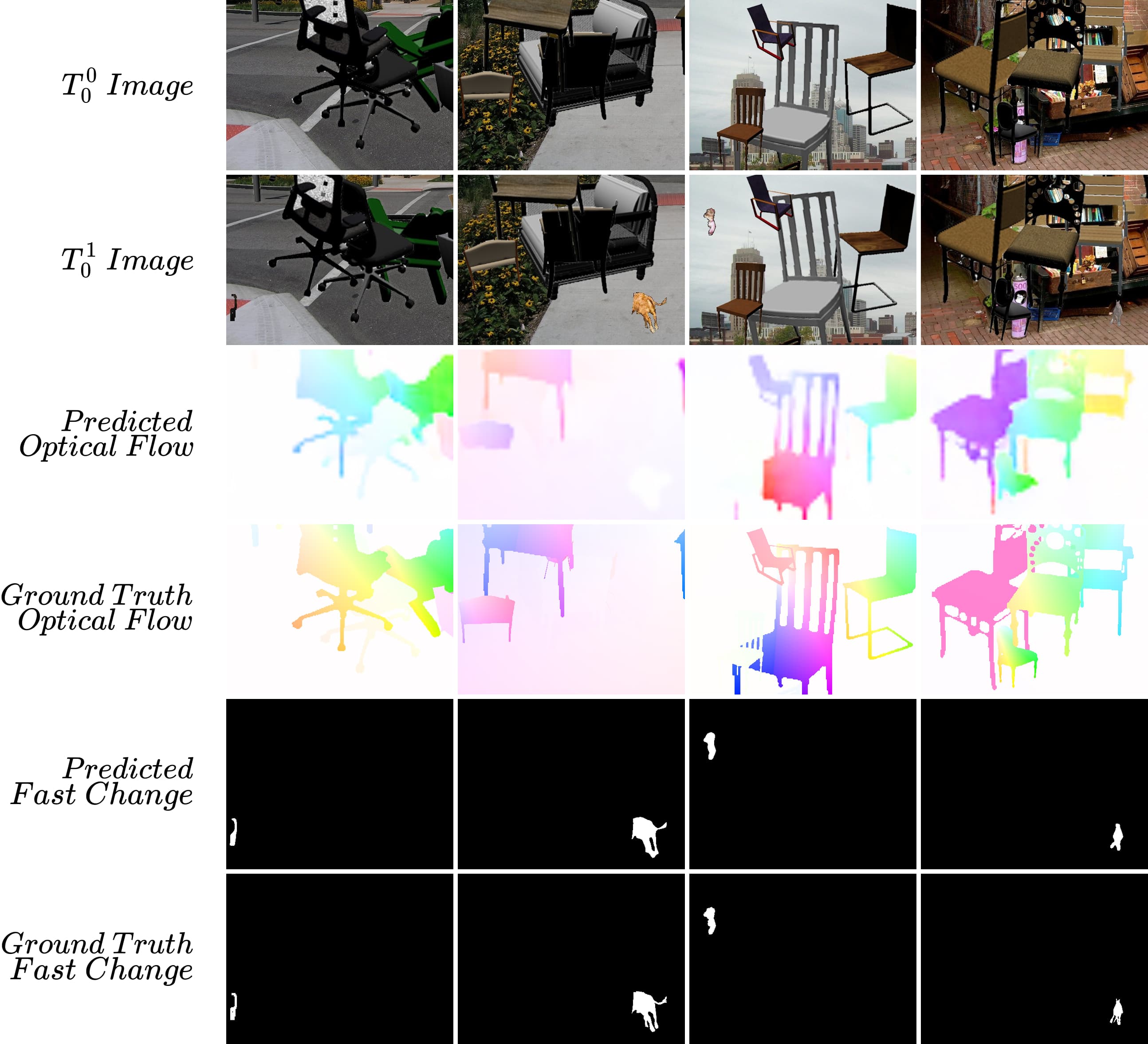}
    \caption{Model result visualization on the Synthetic Flow-Change Dataset}
    \label{fig:vis_dataset}
\end{figure}

\textbf{Visualization.} 
To qualitatively assess the performance of Flow-CDNet on synthetic scenarios with well-controlled dynamic object motion, we utilize visualizations from the Flow Dataset. As shown in Fig.~\ref{fig:vis_dataset}, each row corresponds to a different scene containing foreground fast change objects and varied backgrounds. From left to right, the columns display the bitemporal input images, the ground truth optical flow, and the binary change masks.
In each case, the bitemporal images present noticeable object displacements between two time points, simulating real-world movement such as translation, rotation, or partial occlusion. The ground truth flow maps illustrate the pixel-wise motion fields, where distinct colors represent direction and magnitude of motion, facilitating fine-grained evaluation of optical flow estimation accuracy. Meanwhile, the ground truth change masks (final column) provide supervision for the change detection task by highlighting regions where significant structural changes occur.
This visualization demonstrates our task of making the model’s able to jointly learn both dense motion estimation and sparse change detection, which are essential for handling diverse temporal variations. The Flow Dataset thus serves as a valuable benchmark for validating Flow-CDNet’s capacity to capture object-level dynamics under controlled conditions.

\renewcommand{\dblfloatpagefraction}{.9}

\subsection{Experiment on Real-World Data}

To further evaluate the generalization ability and real-world applicability of Flow-CDNet beyond synthetic benchmarks, we conduct experiments on a curated real-world test set composed of dam bank imagery. This dataset comprises 20 bitemporal image pairs acquired from 12 distinct dam sites, with each pair capturing the same location at two separate time points, denoted as $T_0^0$ and $T_0^1$. The imagery originates from two primary sources: ground-based fixed-position cameras (commonly referred to as trail cameras) and unmanned aerial vehicles (UAVs). The use of trail cameras represents a form of close-range remote sensing, offering continuous, localized monitoring, while UAV imagery provides broader spatial coverage with flexible vantage points. The selected sites reflect a diverse range of real-world change conditions, encompassing both abrupt structural events (e.g., sudden collapses or erosion) and gradual morphological deformations over time. This multi-source dataset enables a comprehensive assessment of the model’s robustness under varied sensing modalities and temporal change dynamics.

Ground truth (GT) annotations were manually constructed to identify regions exhibiting noticeable change between the two time points. However in real-world scenarios, the accurate labeling of fine-grained optical flow vectors presents significant challenges due to complex spatial displacements and the lack of precise reference data. To address this limitation, a unified annotation strategy is adopted, wherein all observable change regions are marked without categorizing them as abrupt or gradual. Consequently, both the detections from the optical flow estimation branch and the change detection branch are considered valid if they correctly localize these annotated regions.

\begin{figure}[!ht]
\centering
\includegraphics[width=\linewidth]{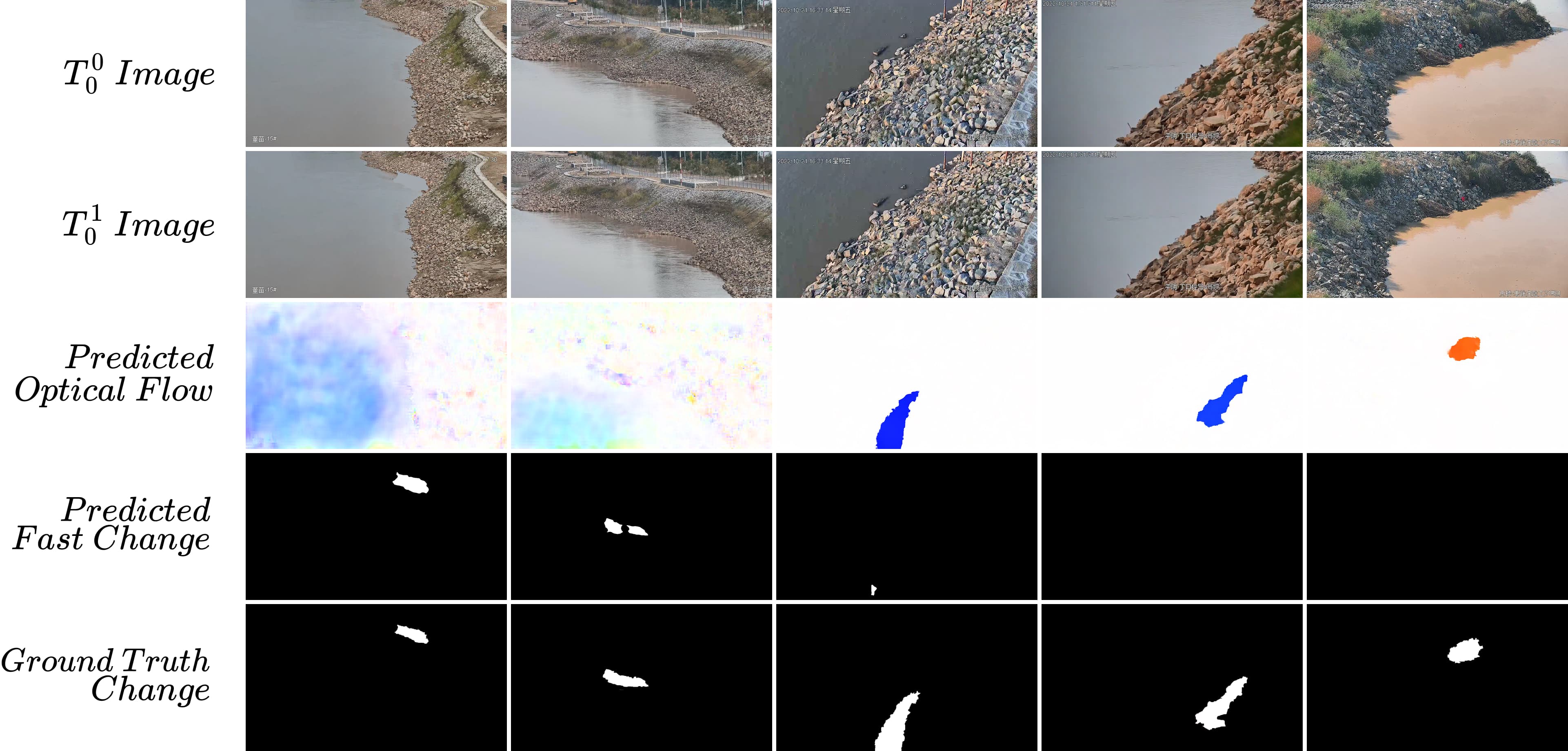}
\caption{Visual comparison of model performance on real-world data. From top to bottom, the rows display: the input image at time $T_0^0$, the input image at time $T_0^1$, the output optical flow estimation from the proposed Flow-CDNet, the output fast change mask from the proposed Flow-CDNet, and the ground truth mask.}
\label{fig:vis_real}
\end{figure}

The model tested in this setting is a pretrained Flow-CDNet, initially optimized on the Flow-Change dataset. During inference, each image pair is processed through the network without any domain-specific fine-tuning, thereby serving as a measure of its robustness to previously unseen data. Representative qualitative results are illustrated in Fig.~\ref{fig:vis_real}. In columns one and two, examples of fast changes are shown, where catastrophic deformation results in substantial structural loss. These are clearly detected by the change detection branch, which effectively captures discrete region-level differences. Columns three to five present examples of slow, progressive surface deformation. Such subtler variations, which may be overlooked by direct difference-based methods, are more effectively revealed through the optical flow estimation branch that encodes temporal pixel-wise displacement information.

This complementary interplay between the two branches—one focusing on binary regional change and the other on continuous motion modeling—underscores the design rationale of Flow-CDNet. Despite the absence of separate GT for fast versus slow changes, the model demonstrates strong performance in capturing both types under a unified evaluation metric. After selecting appropriate change thresholds for different scenes, the final evaluation yields an F1-score of 0.8126, thereby quantitatively affirming the effectiveness of the proposed method. These findings validate the model’s applicability to diverse real-world scenarios and confirm its capability to generalize to complex, uncontrolled environments where both discrete and continuous changes co-occur.

\subsection{Model Efficiency Analysis}

As shown in Table~\ref{tab:model_efficiency}, we evaluate the computational efficiency of three variants of our proposed Flow-CDNet model. The standard Flow-CDNet achieves 14.08 FPS on an RTX 4090 and 12.50 FPS on an RTX 2080, with an inference time of 0.071 seconds and 0.080 seconds, respectively, and a computational cost of 401.8 GFLOPs. The lightweight variants, Flow-CDNet-L and Flow-CDNet-S, offer higher efficiency, reaching 21.73 FPS and 33.33 FPS on an RTX 4090, and 19.23 FPS and 25.32 FPS on an RTX 2080, with reduced computational costs of 215.3 GFLOPs and 212.6 GFLOPs, respectively.

In our downstream task scenario, current inference speed is sufficient to meet application requirements, as real-time performance is not a critical constraint. This allows us to leverage the RAFT-based Flow-CDNet, which prioritizes flow estimation accuracy over computational efficiency compared to lighter backbones like LiteFlowNet and SpyNet. For scenarios demanding higher real-time performance, deploying the model on more powerful hardware can further enhance inference speeds. The reported results across both GPU configurations demonstrate that all variants maintain practical efficiency, with the lightweight models providing significant speed improvements while preserving performance.

\begin{table}[htbp]
\centering
\caption{Model Efficiency Comparison on Different GPUs}
\begin{tabular}{lcccccc}
\toprule
\multirow{2}{*}{Model} & \multirow{2}{*}{FLOPs (G)} & \multicolumn{2}{c}{Inference Time (s)} & \multicolumn{2}{c}{FPS} \\
\cmidrule(r){3-4} \cmidrule(r){5-6}
& & RTX 4090 & RTX 2080 & RTX 4090 & RTX 2080 \\
\midrule
Flow-CDNet   & 401.82 & 0.071 & 0.080 & 14.08 & 12.50 \\
Flow-CDNet-L & 215.30 & 0.046 & 0.052 & 21.73 & 19.23 \\
Flow-CDNet-S & 212.58 & 0.030 & 0.038 & 33.33 & 25.32 \\
\bottomrule
\end{tabular}
\label{tab:model_efficiency}
\end{table}

\subsection{Analysis of Individual Branch Contributions}
\subsubsection{Comparison of Optical Flow Branch Variants}
To ascertain the influence of the optical flow estimation backbone on the overall performance of the Flow-CDNet framework, a qualitative comparison was conducted. This analysis specifically evaluates the generalization capabilities of different model variants when transitioning from a synthetic training environment to real-world application scenarios without any domain-specific fine-tuning. The variants under examination incorporate three distinct optical flow backbones: LiteFlowNet, SpyNet, and RAFT, integrated into our dual-branch architecture as Flow-CDNet-L, Flow-CDNet-S, and the proposed Flow-CDNet, respectively.

\begin{figure}[!htb]
    \centering
    \includegraphics[width=\textwidth]{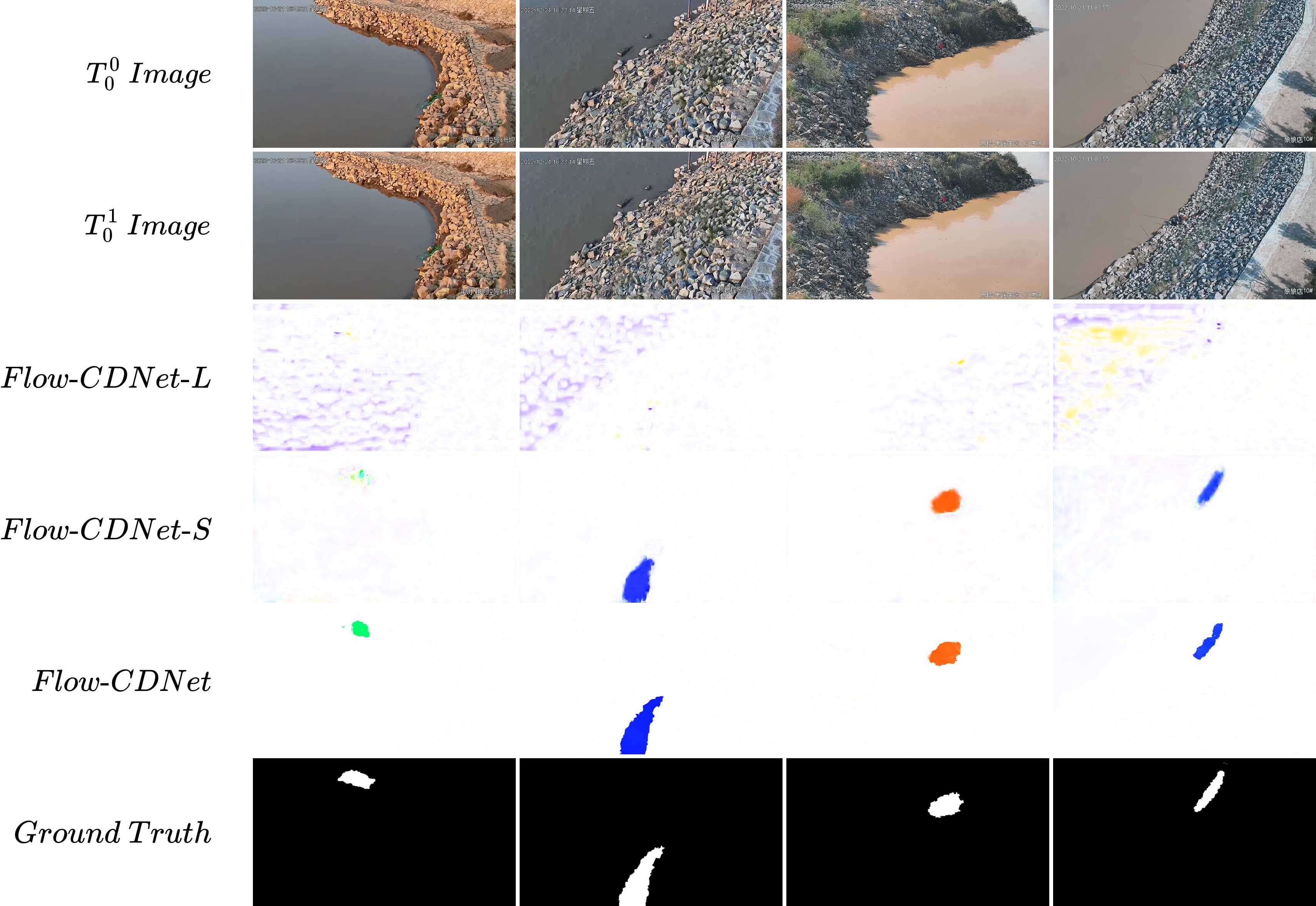}
    \caption{Visual comparison of slow change detection performance using different optical flow backbones on real-world data. From top to bottom, the rows display: the input image at time $T_0^0$, the input image at time $T_0^1$, the output from Flow-CDNet-L (LiteFlowNet), Flow-CDNet-S (SpyNet), the proposed Flow-CDNet (RAFT), and the ground truth mask.}
    \label{fig:of_compare}
\end{figure}

Fig.~\ref{fig:of_compare} illustrates the comparative results on a series of real-world dam bank image pairs, which feature subtle, gradual changes characteristic of slow surface deformation. Each model, having been trained exclusively on the synthetic Flow-Change dataset, was tasked with identifying these changes to assess its cross-domain robustness.

The visual evidence indicates a marked disparity in performance among the backbones. The Flow-CDNet-L variant, which employs LiteFlowNet, struggles significantly with localization. Its output is characterized by diffuse, noisy predictions that fail to form a coherent change mask, indicating that the model's architectural limitations prevent it from generalizing effectively across different data domains without further training.

In contrast, the Flow-CDNet-S model, utilizing SpyNet, demonstrates a moderate improvement. It successfully localizes the general area of change, producing more compact detection masks. However, its performance is compromised by the presence of visual artifacts and a lower degree of shape fidelity when compared to the ground truth. The resulting detections lack the clean boundaries and structural accuracy required for high-fidelity change analysis.

The superior performance of the RAFT-based Flow-CDNet is evident. This configuration consistently generates change masks that are both structurally coherent and spatially precise, aligning closely with the ground truth annotations. The model effectively suppresses noise and preserves the fine-grained boundaries of the deforming regions. This robust generalization from synthetic data to complex, real-world scenes underscores the advanced capabilities of the RAFT architecture in capturing intricate motion cues. The results validate the selection of RAFT as the foundational optical flow branch, as its capacity for accurate and detailed motion estimation is critical to enhancing the change detection accuracy of the unified framework.

\subsubsection{Effect of Flow Estimation on Change Detection Accuracy}
\begin{figure}[!htb]
    \centering
    \includegraphics[width=\textwidth]{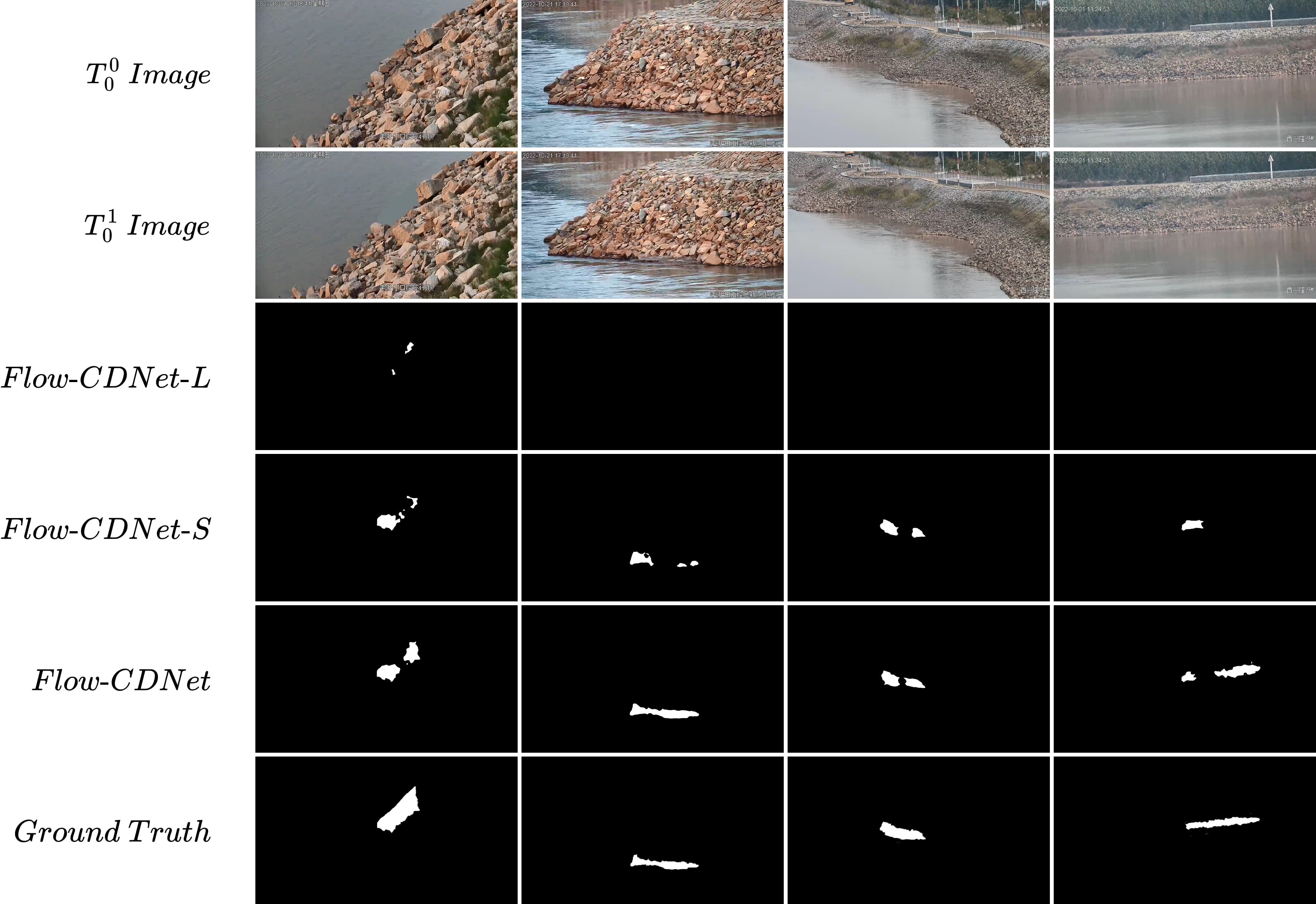}
    \caption{Visual comparison of fast change detection performance with different optical flow backbones on real-world data. From top to bottom, the rows display: the input image at time $T_0^0$, the input image at time $T_0^1$, the output from Flow-CDNet-L (LiteFlowNet), Flow-CDNet-S (SpyNet), the proposed Flow-CDNet (RAFT), and the ground truth mask.}
    \label{fig:cd_compare}
\end{figure}

We examine the influence of the optical flow estimation branch on the performance of the change detection branch. Fig.~\ref{fig:cd_compare} provides a visual comparison of the output from three different Flow-CDNet configurations on real-world data, highlighting their ability to detect rapid changes. Each model was trained exclusively on a synthetic dataset, thereby testing the generalization of the optical flow backbone's estimated motion to drive the change detection task in a new, un-seen domain.

The results reveal a clear dependency between the quality of the estimated optical flow and the precision of the final change mask. The Flow-CDNet-L variant, which incorporates the LiteFlowNet backbone for flow estimation, demonstrates significant limitations. The change detection branch, relying on the low-quality motion information from this backbone, produces fragmented and noisy masks that fail to accurately delineate the changed regions. This indicates that the LiteFlowNet-based flow estimation does not generalize well from synthetic training data, subsequently hampering the downstream change detection task.

In contrast, the Flow-CDNet-S model, which uses SpyNet for optical flow estimation, provides a moderately improved input to the change detection branch. This results in more consolidated detection masks and better localization of changes. However, as depicted in the third and fourth columns, the flow information is still inaccurate, leading to fragmented change masks that miss significant portions of the change region. This illustrates that while the SpyNet-based flow is better than LiteFlowNet's, it is still insufficient to fully support a robust change detection performance.

The proposed Flow-CDNet, leveraging the RAFT architecture for optical flow estimation, consistently provides the most reliable input to the change detection branch. The resulting change masks are structurally coherent, spatially precise, and align closely with the ground truth annotations. The RAFT backbone's superior ability to capture detailed and accurate motion cues from one domain and apply them to another directly translates to enhanced change detection accuracy. The robust generalization of the RAFT-based flow estimation is instrumental in suppressing noise and preserving the fine-grained boundaries of the changing regions, confirming its critical role in boosting the overall performance of the unified framework.

\section{Conclusion}
A novel change detection framework called Flow-CDNet is proposed, which can simultaneously detect slow and fast changes. To train and evaluate this new framework, we build a merged change detection dataset namely Flow-Change, and design a loss function combining binary tversky loss and L2 norm loss, along with a new evaluation metric called FEPE. Evaluations on real-world data indicate that this framework demonstrates robust detection capabilities in authentic scenarios characterized by both gradual and swift changes. Furthermore, through ablation experiments, we verify that the two network branches mutually enhance Flow-CDNet's detection performance. In the future, we'll focus on more challenging scenarios such as defocus, blurriness, or significant lighting variations (day and night).

\bibliographystyle{unsrt}  
\bibliography{references}  






\end{document}